\pdfoutput=1

\documentclass[11pt]{article}

\usepackage{acl}

\usepackage{times}
\usepackage{latexsym}

\usepackage[T1]{fontenc}

\usepackage[utf8]{inputenc}

\usepackage{algorithm}

\usepackage{microtype}
\usepackage{algorithmic}
\usepackage{makecell}
\usepackage{cjhebrew}
\usepackage{amsmath}
\usepackage{graphicx}
\usepackage{url}
\usepackage{multirow}
\usepackage{comment}
\usepackage{amssymb}
\usepackage{pifont}
\usepackage{balance}
\usepackage{mathtools}
\usepackage{xcolor}
\usepackage{colortbl}
\usepackage{bbm}
\usepackage{enumitem}
\usepackage{marvosym}
\usepackage{textcomp}
\usepackage{array}
\usepackage{booktabs}
\usepackage{tabularx}
\usepackage{CJKutf8}

\newcommand{\squishlist}{
\begin{list}{$\bullet$}
{ 
\setlength{\itemsep}{0pt}
\setlength{\parsep}{0pt}
\setlength{\topsep}{0pt}
\setlength{\partopsep}{0pt}
\setlength{\leftmargin}{2em}
\setlength{\labelwidth}{1.5em}
\setlength{\labelsep}{0.5em} } }
\definecolor{cadmiumgreen}{rgb}{0.0, 0.42, 0.24}
\newcommand{\squishend}{
\end{list} }

\title{Character-level White-Box Adversarial Attacks against Transformers via  Attachable Subwords Substitution}

\author{Aiwei Liu, Honghai Yu, Xuming Hu, Shu'ang Li, Li Lin, Fukun Ma, \\ \textbf{ Yawen Yang, Lijie Wen$^{\dagger}$}\\
  Tsinghua University\\  
  \small \texttt{\{liuaw20, yhh21, hxm19, lisa18, lin-l16, mafk19, yyw19\}@mails.tsinghua.edu.cn}\\
  \small \texttt{wenlj@tsinghua.edu.cn}
   \thanks{$^\dagger$Corresponding author.}
  }

\begin{document}
\maketitle

\newcommand{\modelname}{\texttt{CWBA}}
\begin{abstract}
We propose the first character-level white-box adversarial attack method against transformer models. The intuition of our method comes from the observation that words are split into subtokens before being fed into the transformer models and the substitution between two close subtokens has a similar effect to the character modification. Our method mainly contains three steps.  First, a gradient-based method is adopted to find the most vulnerable words in the sentence. Then we split the selected words into subtokens to replace the origin tokenization result from the transformer tokenizer. Finally, we utilize an adversarial loss to guide the substitution of attachable subtokens in which the Gumbel-softmax trick is introduced to ensure gradient propagation.
Meanwhile, we introduce the visual and length constraint in the optimization process to achieve minimum character modifications.
Extensive experiments on both sentence-level and token-level tasks demonstrate that our method could outperform the previous attack methods in terms of success rate and edit distance. Furthermore,  human evaluation verifies our adversarial examples could preserve their origin labels.

\end{abstract}

\section{Introduction}

Adversarial examples are modified input data that could fool the machine learning models but not humans.  Recently,  Transformer \cite{vaswani2017attention} based model such as BERT \cite{devlin-etal-2019-bert} has achieved dominant performance on a wide range of natural language process (NLP) tasks. Unfortunately, many works have shown that transformer-based models are  vulnerable to adversarial attacks  \cite{guo-etal-2021-gradient, garg-ramakrishnan-2020-bae}. On the other hand, the adversarial attack could help improve the robustness of models through adversarial training, which emphasizes the importance of finding high-quality adversarial examples.

\begin{figure}
  \includegraphics[width=0.47\textwidth]{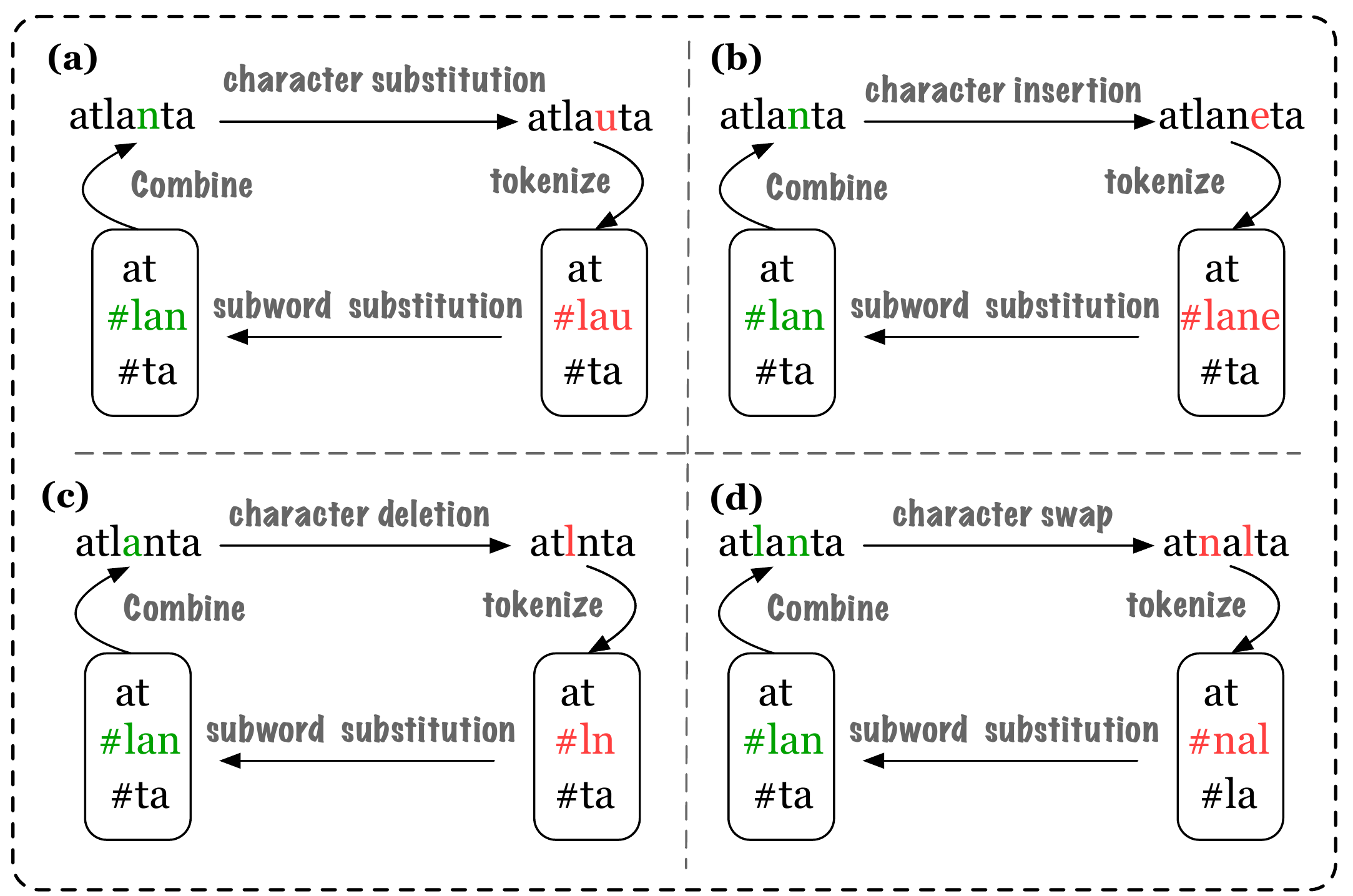}
  \caption{Subtoken substitution operation could achieve the same result as all four character modification operations.}
  \label{fig:intro}
  \vspace{-0.2in}
\end{figure}

Recently, some efficient and effective  attacking methods have been proposed at token level  (e.g. synonym substitution) \cite{guo-etal-2021-gradient} and sentence level  (e.g.  paraphrasing  input  texts) \cite{wang-etal-2020-t3}.   However, this is not the case in character-level attack methods (e.g. mistyping words), which barely hinder human understanding and is thus a natural attack scenario.  Most  previous methods \cite{gao2018black, eger-benz-2020-hero} achieve the character-level attack in a black box manner,  which requires hundreds of attempts and the attack success rate is not good enough. White box attack methods are natural solutions to these drawbacks, but current character-level white box attack methods \cite{ebrahimi-etal-2018-hotflip, ebrahimi-etal-2018-adversarial} only work for models taking characters as input and thus fail on  token-level transformer model.

Achieving  character-level white box attack via single character modification is impossible for the transformer model, due to the  gradient of characters being unavailable.  We choose to implement the character-level attack via subtoken substitution based on the following two observations. (1) Nearly all transformer-based pre-training models adopt subword tokenizer \cite{sennrich-etal-2016-neural}, in which each word is split into subtokens containing one start subtoken and several subtoken attached to it (attachable subtoken). (2) As shown in Figure \ref{fig:intro},  all character modifications (e.g. swap and insertion) can be achieved by subtoken substitution.   


Based on the above observations, we propose {\modelname},  the first \textbf{C}haracter-level \textbf{W}hite-\textbf{Box} \textbf{A}ttack method against transformer models via attachable subwords substitution.   Our method mainly contains three steps: target word selection, adversarial tokenization, and subtoken search. 

Since our {\modelname} requires specific words as input, finding the most vulnerable words is required. Our model first ranks the words according to the gradient of words from our adversarial goal. Then during the adversarial tokenization process,  the top-ranked words are split into at least three subtokens, including a start subtoken and several attachable subtokens. Our {\modelname} method aims to replace these attachable subtokens to achieve character attack. 

Due to the discrete nature of natural languages prohibits the gradient optimization of subtokens, we leverage the Gumbel-Softmax trick \cite{DBLP:conf/iclr/JangGP17} to sample a continuous distribution from tokens and thus allow gradient propagation. The attachable subtokens are then optimized by a gradient descent method to generate the adversarial example. Meanwhile, to minimize the degree of modification, we also introduce visual and length constraints during optimization to make the replaced subtokens visually and length-wise similar.

Our {\modelname} method could outperform previous attack methods on both sentence level (e.g. sentence classification) and token level (e.g. named entity recognition) tasks in terms of success rate and edit distance. It is worth mentioning that {\modelname} is the first white box attack method applied to token-level tasks. Meanwhile, we demonstrate the effectiveness of {\modelname} against various transformer-based models.  Human evaluation experiments verify our adversarial attack method is label-preserving. Finally, the adversarial training experiment shows that training with our adversarial examples would increase the robustness of models. 

To summarize, the main contributions of our paper are as follows:
\begin{itemize}
    \item To the best of our knowledge, {\modelname} is the first character-level white box attack method against transformer models.
    \item Our {\modelname} method is also the first white box attack method applied to token-level tasks.
    \item We propose a visual constraint to make the replaced subtoken similar to the original one.
    \item Our {\modelname} method could outperform the previous attack methods on both sentence-level tasks and token-level tasks. \footnote{Code and data are available at  \href{https://github.com/THU-BPM/CWBA}{https://github.com/THU-BPM/CWBA} } 
\end{itemize}
\section{Related Work}

\subsection{White box attack method in NLP}

White box attack methods could find the defects of the model with low query number and high success rate,  which have been successfully applied to image and speech data \cite{DBLP:conf/iclr/MadryMSTV18, DBLP:conf/sp/Carlini018}. However,  applying white-box attack methods to natural language is more challenging due to the discrete nature of the text.  To search the text under the guidance of gradient and achieve a high success rate, \citet{cheng-etal-2019-robust, cheng-etal-2019-evaluating} choose to optimize in the embedding space and search the nearest word, which suffers from high bias problems.  To further reduce the bias, \citet{cheng2020seq2sick} and \citet{DBLP:conf/ijcai/SatoSS018} restrict the optimization direction towards the existing word embeddings.  However, the optimization process of these methods is unstable due to the sparsity of the word embedding space.  Other methods try to directly optimize the text by gradient estimation techniques such as Gumbel-Softmax sampling \cite{xu-etal-2021-grey, guo-etal-2021-gradient}, reinforcement learning \cite{zou-etal-2020-reinforced},  metropolis-hastings sampling \cite{zhang-etal-2019-generating-fluent}.  Our {\modelname} adopts the Gumbel-Softmax technique for subtokens to achieve the character-level white-box attack.


\subsection{Attack method against Transformers}

Transformer-based \cite{vaswani2017attention} pre-training models \cite{devlin-etal-2019-bert, liu2019roberta} have shown
their great advantage on various NLP tasks.  However, recent works reveal that these pretraining models are vulnerable to adversarial attacks under many scenarios such as sentence classification \cite{li-etal-2020-bert-attack}, machine translation \cite{cheng-etal-2019-robust}, text entailment \cite{xu2020rewriting} and part-of-speech tagging
\cite{eger-benz-2020-hero}. Most of these methods achieve attack in the black box manner, which are implemented by character modification \cite{eger-benz-2020-hero}, token substitution \cite{li-etal-2020-bert-attack} or sentence paraphrasing \cite{xu2020rewriting}.
However, these black-box attack methods usually require hundreds of queries to the target model and the success rate cannot be guaranteed. To alleviate these problems, some white-box attack methods have been proposed including token-level methods \cite{guo-etal-2021-gradient} and sentence-level methods \cite{wang-etal-2020-t3}. Different from these methods, our {\modelname} is the first character-level white-box attack method for transformer-based models.
\section{Methods}


In this section, we detail our proposed framework {\modelname} for the character-level white-box attack method. In the following content, we first give a formulation of our attack problem, followed by a detailed description of the three key components: target word selection, adversarial tokenization, and subtoken search.

\subsection{Attack Problem Formulation}

We formulate the adversarial examples as follows.  Given an input sentence $\mathbf{x} = (x_{1}, x_{2}, ... , x_{n})$ with length $|n|$, suppose the classification model $\mathbf{H}$ could predict the correct corresponding sentence or token label $y$ such that $\mathbf{H}(\mathbf{x}) = y$.  An adversarial example is a sample $\mathbf{x'}$ close to $\mathbf{x}$ but causing different model prediction such that  $\mathbf{H}(\mathbf{x'}) \neq y$.

The process of finding adversarial examples is modeled as a gradient optimization problem. Specifically, given the classification logits vector $\mathbf{p} \in \mathbb{R}^{K} $ generated by model $\mathbf{H}$ with $K$ classes, the adversarial loss is defined as the margin loss:
\begin{equation}
\begin{aligned}
\ell_{\operatorname{adv}}(\mathbf{x}, y)= 
 \max \left(\mathbf{p}_{y}- \max _{k \neq y} \mathbf{p}_{k} + \kappa, 0\right),
\end{aligned}
\label{eq:eq1}
\end{equation}
which motivates the model to misclassify $\mathbf{x}$ by a margin $\kappa > 0$. The effectiveness of margin loss has been validated in many attack algorithms \cite{guo-etal-2021-gradient, DBLP:conf/sp/Carlini018}.

Given the adversarial loss $\ell_{\operatorname{adv}}$, the goal of our attack algorithm can be modeled as a constrained optimization problem:
\begin{equation}
\min \ell_{\operatorname{adv}}\left(\mathbf{x}^{\prime}, y\right) \quad \text { subject to } \rho\left(\mathbf{x}, \mathbf{x}^{\prime}\right) \leq \epsilon,
\label{eq:goal}
\end{equation}
where $\rho$ is the function measuring the similarity between origin and adversarial examples. In our work,  the similarity is measured using the edit distance metric \cite{DBLP:journals/pami/YujianB07}.

\subsection{Target Word Selection}
\label{sec:select}

Since our attack method takes specific words as the target and performs pre-processing to these words, obtaining the most critical words for target task prediction is required.  To find the most vulnerable words, we sort the words based on the l2 norm value of gradient towards adversarial loss in Eq \ref{eq:eq1}:
\begin{equation}
    \hat{\mathbf{x}} = \underset{\mathbf{x}}{\operatorname*{argsort}}\left( \|\nabla_{x_1} \ell_{\operatorname{adv}}\|_{2}, ..., \|\nabla_{x_{n}} \ell_{\operatorname{adv}}\|_{2} \right)
\end{equation}
where  $\nabla_{x_j} \ell_{\operatorname{adv}}$ is the gradient of the $j$-th token. Note that word $x_j$ may be tokenized into several subtokens $[t_{j0}...t_{jn}]$, and its gradient is defined as the average gradient of these subtokens:
\begin{equation}
     \|\nabla_{x_j} \ell\|_{2} =  \operatorname{avg}\left( \|\nabla_{t_{j0}} \ell\|_{2},...,  \|\nabla_{t_{jn}} \ell\|_{2} \right),
\end{equation}
 where the loss $\ell$ is the adversarial loss $\ell_{\operatorname{adv}}$ in our work.
 Our {\modelname} would take the first $N$ words from the sorted word list  $\hat{\mathbf{x}}$ as targets, where $N$ is a task-related hyperparameter.

\subsection{Adversarial Tokenization}
\label{sec:tokenize}


The selected words are required to split into subtokens before performing the character-level attack. We observe that the transformer tokenizer has the following two properties: (1) The correctly spelled words usually won't split or only split into a few subtokens. (2) The misspelled words are tokenized into more subtokens than the correctly spelled words. For example, the word \textit{boston} won't be segmented but after single character modification,  \textit{bosfon} would be tokenized into three subtokens \textit{bo}, \textit{\#sf} and \textit{\#on}.  To keep the tokenization consistency during the attack, we propose the adversarial tokenizer which tokenizes the correctly spelled words into more subtokens than the transformer tokenizer.

To further improve the tokenization consistency during the attack process, our main principle is to make the subtokens as long as possible, since longer subtokens are more difficult to combine with characters to form new subtokens\footnote{More details and statistics are provided in the appendix}. Specifically, our tokenization  contains the following steps:
\squishlist
    \item[\textbf{1.}] Find the longest subwords in the first half of the word to form the longest start subtoken.
    \item[\textbf{2.}] Find the longest subwords in the second half of the word to form the longest end subtoken.
    \item[\textbf{3.}] Tokenize the rest part with the transformer tokenizer to generate the middle subtokens.
\squishend
After these steps, we obtain the longest start and end subtokens and our algorithm would substitute the middle subtokens, which keeps the maximum consistency of tokenization during the attack.
\begin{figure}
\centering
  \includegraphics[width=0.47\textwidth]{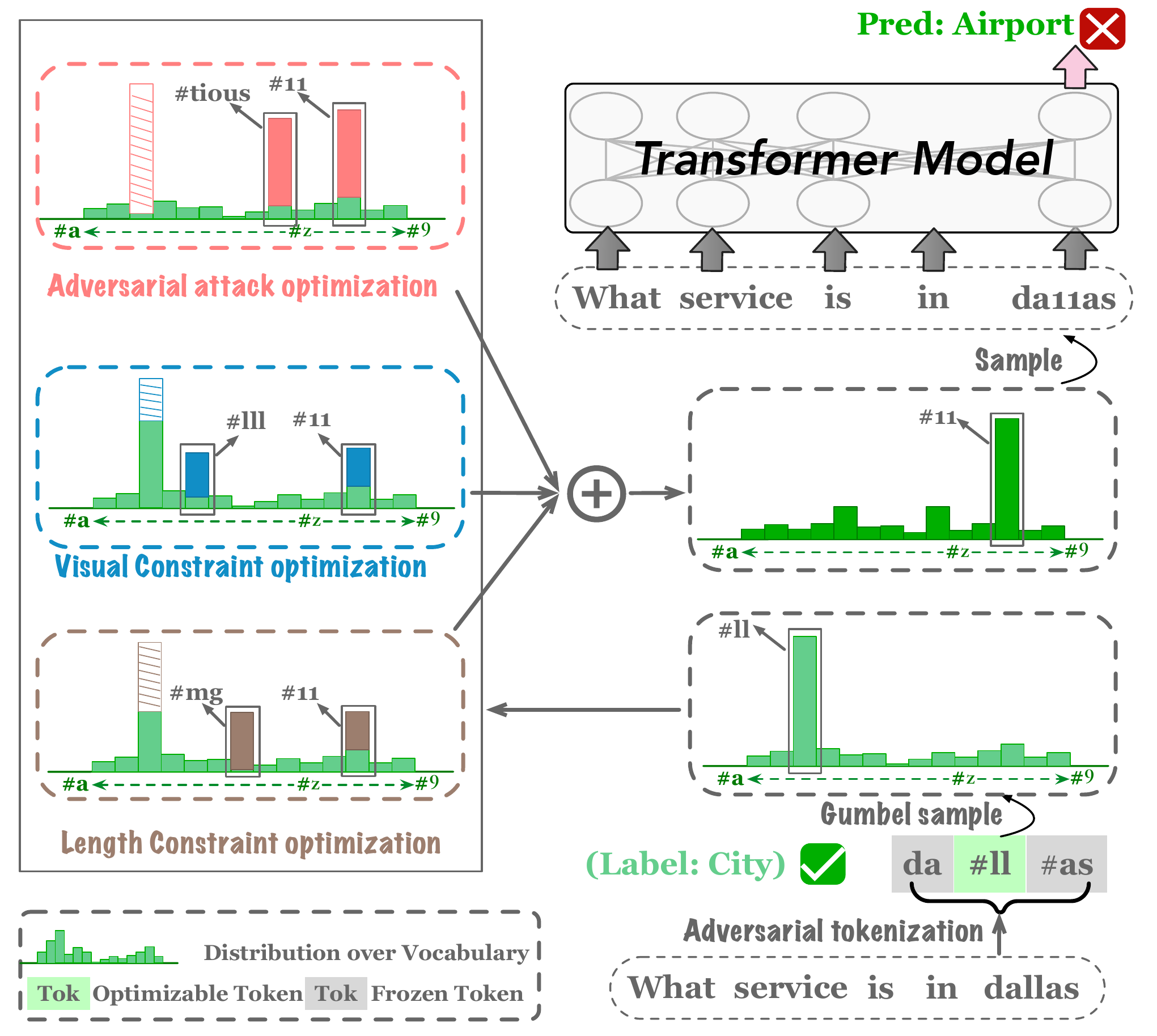}
  \caption{Details of the subtoken search module. }
  \label{fig:method}
  \vspace{-0.1in}
\end{figure}

\subsection{Subtoken Search}
\label{sec:optimization}

After obtaining the vulnerable words and tokenizing them into subtokens in an adversarial way, the subtoken search module aims to find new subtokens for substitution to construct adversarial examples. 

As shown in Figure \ref{fig:method}, to allow gradient propagation, the target subtoken is first transformed from the discrete distribution to a continuous distribution by the Gumbel-softmax trick \cite{DBLP:conf/iclr/JangGP17}.  Then the continuous distribution is optimized by three objectives: adversarial attack, visual constraint, and length constraint to search the adversarial examples with minimal modifications. The final adversarial examples could be sampled from the optimized Gumbel-softmax distribution.

\noindent\textbf{Computing  gradients using Gumbel-softmax}\\
Since the origin subtoken input is represented in the discrete categorical distribution over vocabulary, the gradient could not be propagated directly. We adopt the Gumbel-softmax approximation to derive the soft estimation of the gradient.

Specifically, for any token $x_{i} \in \mathcal{V}$ from a fixed vocabulary $\mathcal{V} = \{1,..., V\}$, we denote its one-hot distribution as $\phi_{i}$. The Gumbel-softmax distribution $\pi_i$ could be represented as follows:
\begin{equation}
\left(\pi_{i}\right)_{j} =\frac{\exp \left(\left(\phi_{i, j}+g_{i, j}\right) / T\right)}{\sum_{v=1}^{V} \exp \left(\left(\phi_{i, v}+g_{i, v}\right) / T\right)}, 
\label{eq:gumbel}
\end{equation}
where $j$ indicates the jth token in the dictionary, $g_{i, j}$ is sampled from the uniform distribution $U(0,1)$ to introduce randomness and $T$ is the temperature parameter of the Gumbel-softmax distribution.  $\phi_{i}$ could be updated by gradient through the Gumbel-softmax estimation  $\pi_i$.  Let $\mathbf{e}$ be the embedding lookup table of the transformer, the embedding vector of the distribution $\pi_i$ can be defined as:
\begin{equation}
\mathbf{e}\left(\pi_{i}\right)=\sum_{j=1}^{V}\left(\pi_{i}\right)_{j} \mathbf{e}(j),
\label{eq:emb}
\end{equation}
which is the input to the transformer model.

\noindent\textbf{Adversarial attack objective}\\
To search for the desired subtoken substitution which could mislead the model, an effective objective function is required. In practice, We adopt the margin loss in Eq \ref{eq:eq1}. Given the whole sentence vector from Gumbel-softmax distribution $\mathbf{e}(\boldsymbol{\pi}) =\mathbf{e}\left(\pi_{1}\right), ..., \mathbf{e}\left(\pi_{n}\right)$, the adversarial loss is represented as  $\ell_{\operatorname{adv}}\left(\mathbf{e}(\boldsymbol{\pi}), y\right)$. Note that the Gumbel-softmax distribution is only applied on the target subtokens while the discrete distribution keeps unchanged on other subtokens such that $\pi_i = \phi_i$. For example, only the subword \textit{\#ll} in Figure \ref{fig:method} is sampled by Gumbel-softmax method while the subtokens \textit{da} and \textit{\#as} maintains one-hot distribution.

However, the attack success of continuous distribution $\pi_i$ does not guarantee the top one probability tokens in $\pi_i$ could fool the target model. To reduce the gap between the distribution $\boldsymbol{\pi}$ and one-hot word distribution, we sample a discrete distribution from $\boldsymbol{\pi}$ and the adversarial loss is also applied on it, which is defined as follows:
\begin{equation}
(\hat{\pi}_{i})_{j}=\left\{\begin{aligned}
\frac{(\pi_{i})_{j}}{ |(\pi_{i})_{j}|} , \quad & (\pi_{i})_{j} = \operatorname{max}\left(\pi_{i}\right) \\
0,\quad &  (\pi_{i})_{j} \neq \operatorname{max}\left(\pi_{i}\right)
\end{aligned}\right..
\label{eq:implict}
\end{equation}
The distribution  $\boldsymbol{\hat{\pi}}$ is the one-hot distribution of the previous top probability tokens, of which the gradient is retained. Finally, our adversarial loss could be represented as:
\begin{equation}
\ell_{\operatorname{adv}} = \ell_{\operatorname{adv}}\left(\mathbf{e}(\boldsymbol{\pi}), y\right) + \lambda_{adv} \ell_{\operatorname{adv}}\left(\mathbf{e}(\boldsymbol{\hat{\pi}}), y\right),
\label{eq:adv}
\end{equation}
where the first term could quickly explore the candidate tokens and the second term further exploits the attack effect of the top one probability token.  $\lambda_{adv}$ is a hyper-parameter that balances the trade-off of exploration and exploitation.

\noindent\textbf{Visual constraint objective}\\
To minimize the edit distance (Eq. \ref{eq:goal}) caused by subtoken substitution, the visual constraint restricts the substituted subtoken visually similar to the original one, such as the \textit{\#ll} and \textit{\#11} in Figure \ref{fig:method}.

In practice, we first generate the images of all subtokens in \textit{helvetica} font.  Then the pre-trained ResNet50 network \cite{he2016deep} is adopted to transform all the token images into vectors.  Let $\mathbf{v}(i)$ be the visual embedding of $i$-th token in the vocabulary. Similar to Eq. \ref{eq:emb},  the visual embedding of distribution $\pi_i$ could be represented as follows:
\begin{equation}
\mathbf{v}\left(\pi_{i}\right)=\sum_{j=1}^{V}\left(\pi_{i}\right)_{j} \mathbf{v}(j).
\end{equation}
The visual constraint aims to minimize the gap of visual embedding between origin subtoken $x_i$ and distribution $\pi_i$, which is defined as:
\begin{equation}
    \ell_{\operatorname{vis}} = \sum_{I} \| \mathbf{v}(\pi_i) - \mathbf{v}(x_i)  \|_{2}, 
\end{equation}
where $I$ is the set of all replaceable subtokens and $\| \|_{2}$ is the l2 normalization operation.

\noindent\textbf{Length constraint objective}\\
To further reduce character modifications,  the length constraint objective aims to keep the length of subtoken unchanged during the attack process.

Similar to the visual constraint objective, the length of the distribution $\pi_i$ could be defined as: 
\begin{equation}
\mathbf{l}\left(\pi_{i}\right)=\sum_{j=1}^{V}\left(\pi_{i}\right)_{j} \mathbf{l}(j),
\end{equation}
where $\mathbf{l}(i)$ is the length of the i-th token.  And the length constraint loss could be represented similarly to the visual constraint loss:
\begin{equation}
    \ell_{\operatorname{len}} = \sum_{I} \| \mathbf{l}(\pi_i) - \mathbf{l}(x_i)  \|_{2}.
\end{equation}
\noindent\textbf{Objective function}\\
Our final objective is the combination of adversarial attack objective, visual constraint objective, and length constraint objective:
\begin{equation}
\mathcal{L} = \ell_{\operatorname{adv}} + \lambda_{vis}\ell_{\operatorname{vis}} + \lambda_{len}\ell_{\operatorname{len}},
\label{eq:loss}
\end{equation}
where $\lambda_{vis}, \lambda_{len} > 0$ are hyperparameters that controls the degree of constraints.  The final loss $\mathcal{L}$ is minimized using the gradient descent method.

 Note that the number of attacked words is difficult to set in long sentences. So we search between two hyperparameters $N_1$ and $N_2$ until the adversarial loss could be well optimized.  The process of our algorithm is summarized in algorithm \ref{alg:attack}.

\begin{algorithm}[h]
\algsetup{linenosize=\small}
\caption{CWBA Attack}\label{alg:attack}
\footnotesize
\begin{algorithmic}[1]
\STATE Input words: $\mathbf{x}=(x_0, ..., x_n)$,  label: $y$
\STATE Get sorted word list $\mathbf{\hat{x}} = [x_{top-1}, x_{top-2},...]$ from Eq \ref{eq:eq1} based on the importance of words
\FOR {$k=k_1$ to $k_2$} 
 \STATE $\mathbf{s} = topk(\mathbf{\hat{x}})$  {\color{cadmiumgreen} // Get the most important k words} 
 \STATE $\textbf{h} \gets []$ {\color{cadmiumgreen} //  Input token distribution } 
 \FOR {$x_i \in \textbf{x}$}
 \IF{$x_i \in \textbf{s}$}
 \STATE  $[t_0, t_1 ..., t_n] = \operatorname{adv\_tokenize}(x_i)$ (Sec \ref{sec:tokenize})
 \STATE  $[\phi_0, \phi_1 ..., \phi_n] = \operatorname{Onehot}([t_0, t_1 ..., t_n])$
 \STATE  $[\pi_1, ., \pi_{n-1}] = \operatorname{Gumbel}([\phi_1, ., \phi_{n-1}] )$ (Eq \ref{eq:gumbel})
 \STATE  {\color{cadmiumgreen} //  Only search the middle subtokens} 
 \STATE  $\mathbf{h} = \mathbf{h} \cup [\phi_0] \cup [\pi_1, ..., \pi_{n-1}] \cup [\phi_{n}]$
  \ELSE
  \STATE $[t_0, t_1 ..., t_n] = \operatorname{transformer\_tokenize}(x_i)$ 
  \STATE $[\phi_0, \phi_1 ..., \phi_n] = \operatorname{Onehot}([t_0, t_1 ..., t_n])$
  \STATE $\mathbf{h} = \mathbf{h} \cup [\phi_0, ..., \phi_{n}]$
 \ENDIF
 \ENDFOR
 \FOR {$i=0$ to $\operatorname{MAX\_ITER}$} 
 \STATE Get loss $\mathcal{L}$ from Eq \ref{eq:loss} based on input $\textbf{h}$
 \STATE Update $\textbf{h}$ using gradient descent method
 \ENDFOR
 \STATE Get adversarial loss $\ell_{\operatorname{adv}}$ from Eq \ref{eq:adv}
 \STATE {\color{cadmiumgreen} // Whether Adversarial loss is well optimized } 
 \IF{$\ell_{\operatorname{adv}} < \kappa$} 
 \STATE Jump to line 3 {\color{cadmiumgreen} // Attack fail, search more words } 
 \ENDIF
 \STATE Sample sentence $\mathbf{\tilde{x}}$ from $\textbf{h}$
  \IF{$f(\mathbf{\tilde{x}}) \neq y$}
  \STATE {\color{cadmiumgreen} // Attack success } 
  \ENDIF
\ENDFOR
\RETURN $\mathbf{\tilde{x}}$

\end{algorithmic}
\end{algorithm}
\section{Experiments}
We conduct extensive experiments on eight datasets across two tasks (sentence classification and token classification) and four transformer models (BERT, RoBERTa, XLNet, and ALBERT) to show the effectiveness of {\modelname} on white-box attack scenarios and give a detailed analysis to show its advantage.

\subsection{Experiment Setup} 

\noindent\textbf{Datasets.}  The sentence classification datasets include \textbf{DBPedia} \cite{DBLP:journals/semweb/LehmannIJJKMHMK15}, \textbf{AG News} \cite{NIPS2015_250cf8b5} for article/news categorization and \textbf{Yelp Reviews}\cite{NIPS2015_250cf8b5}, \textbf{IMDB} \cite{DBLP:conf/acl/MaasDPHNP11} for sentiment classification. And the token classification datasets include \textbf{ATIS} \cite{DBLP:conf/slt/TurHH10}, \textbf{SNIPS}\footnote{\href{https://github.com/sonos/nlu-benchmark}{https://github.com/sonos/nlu-benchmark}} for slot filling and \textbf{CONLL-2003} \cite{tjong-kim-sang-de-meulder-2003-introduction}, \textbf{Ontonotes} \cite{pradhan-etal-2013-towards} for named entity recognition (NER). \\

\noindent\textbf{Baselines.} For the sentence classification task, we adopt five competitive baselines which are various in attack settings. \textbf{GBDA} \cite{guo-etal-2021-gradient} is a white-box attack method which performs token replacement under the gradient guidance.  \textbf{BERT-Attack} \cite{li-etal-2020-bert-attack}, \textbf{BAE} \cite{garg-ramakrishnan-2020-bae} and \textbf{TextFooler} \cite{jin2020bert} aim to replace tokens in the black-box manner.  \textbf{DeepWordBug} \cite{gao2018black} is a black-box attack method which modifies characters of the most vulnerable words.  Note that the edit distance of adversarial examples generated by token replacement is much higher than that of character modification.

For the token classification task, we adopt two baselines. \textbf{Zéroe} \cite{eger-benz-2020-hero} explores several character-level black-box attack methods of which we choose the vision method as our baseline for its excellent attack effect. 
\textbf{DeepWordBug} is also adopted as a competitive baseline, which modifies the characters of keywords (e.g. entity in named entity recognition).

More details about these baselines are provided in the appendix. \\

\noindent\textbf{Hyper-parameters.} The input token distribution is optimized by Adam \cite{DBLP:journals/corr/KingmaB14} with a learning rate of $0.3$ for $100$ iterations (token classification) or $300$ iterations (sentence classification). The margin $\kappa$ of adversarial loss is set to 7 for sentence classification and 5 for token classification.  And the $T$ of Gumbel-Sampling (Eq. \ref{eq:gumbel}) is set to 1.  The loss weights $\lambda_{adv}$, $\lambda_{vis}$ and $\lambda_{len}$ (Eq.\ref{eq:loss}) are set to 1, 0.1, 2 respectively.  The $N_1$ and $N_2$ are set to 2, 15 and 1, 2 for sentence classification and token classification tasks respectively. \\

\noindent\textbf{Models.} We attack four transformer models with our {\modelname} method: BERT \cite{devlin-etal-2019-bert}, XLNet \cite{yang2019xlnet}, RoBERTa \cite{liu2019roberta} and ALBERT \cite{DBLP:conf/iclr/LanCGGSS20}. We first fine-tune these models on target datasets and then perform the attack. All of these models utilize the subword tokenization method. The BERT tokenizer adds a prefix \#\# to the attachable subwords. The tokenizer of RoBERTa adds a prefix Ġ to the start subwords. And the tokenizer of XLNet and ALBERT adds a prefix \textbf{\_} to the start subtokens.
\subsection{Quantitative Evaluation.}

\noindent\textbf{Sentence-level attacks.} Table \ref{tab:sentence} shows the attack performance on the sentence classification task with the BERT classifier. Following the previous works \cite{guo-etal-2021-gradient}, we randomly select 1000 inputs from the test set as attack targets.  Our method searches the number of attacked words between $N_1$ and $N_2$ until the attack succeeds. The adversarial accuracy (Adv.Acc.) is the accuracy of the last searched examples. The Edit Dist represents the sum of edit distances for all modified words.



\begin{table}[bt!]
\centering
\resizebox{0.9\linewidth}{!}{
\begin{tabular}{lccccccc}
\toprule
\textbf{Datasets} & \textbf{Clean Acc.}  & \textbf{Attack Alg.} & \textbf{Adv.Acc.}  & \textbf{\#Queries} & \textbf{Edit Dist}    \\
\midrule 
\multirow{ 5}{*}{\textbf{AG News}} & \multirow{ 5}{*}{95.1}  & CWBA(ours) & \textbf{3.2} & 6.1 & \textbf{17.3} \\
 & & DeepWordBug & 23.7 & 319 & 23.4 \\
 & & GBDA & 3.5 & \textbf{5.8}  & 76.3\\
  & & BERT-Attack & 10.6 & 213  & 83.4 \\
 & &  BAE & 13.0 & 419 & 65.2 \\
 & & TextFooler & 12.6  & 357& 97.5 \\
\midrule 
\multirow{5}{*}{\textbf{Yelp}} & \multirow{ 5}{*}{97.3}  &CWBA(ours) & \textbf{4.0} & \textbf{7.2} &\textbf{18.5}\\
 & &DeepWordBug & 27.7 & 543 & 19.8 \\
 & &GBDA & 4.4 & 7.6 & 84.1\\
 & & BERT-Attack & 5.1 & 273  & 95.3\\
 & &BAE & 12.0 & 434 & 63.1 \\
 & &TextFooler & 6.6 & 743 & 74.6 \\

 \midrule 
 \multirow{ 5}{*}{\textbf{IMDB}} & \multirow{ 5}{*}{93.0} & CWBA(ours) & \textbf{5.4} & 8.5 & \textbf{23.4}\\
 & &DeepWordBug & 28.4 & 134& 24.5 \\
 & &GBDA & 5.6 & \textbf{5.2} & 84.5\\
 & & BERT-Attack & 11.4 & 454  & 91.3\\
 & &BAE &24.0 & 592 & 88.2\\
 & &TextFooler &13.6& 1134 & 77.1 \\

 \midrule 
  \multirow{ 5}{*}{\textbf{DBPedia}} & \multirow{ 5}{*}{99.2} & CWBA(ours) & \textbf{6.9} &\textbf{5.3} & \textbf{19.1}\\
 & &DeepWordBug & 19.4 & 453 & 34.5 \\
 & &GBDA & 7.1 & 5.6 & 79.1 \\
 & &BERT-Attack & 8.5 & 487 & 86.5\\
 & &BAE & 10.4 & 398 & 94.3\\
 & &TextFooler &9.5 & 829 & 83.4\\

\bottomrule
\end{tabular}}
\vspace{-1mm}
\caption{Attack result on sentence classification datasets with finetuned BERT classifiers. }
\label{tab:sentence}
\end{table}



\begin{table}[bt!]
\centering
\resizebox{\linewidth}{!}{
\begin{tabular}{lcccccc}
\toprule
\textbf{Datasets} & \textbf{Clean F1.}  & \textbf{Attack Alg.} & \textbf{Adv.F1.} & \textbf{success rate} & \textbf{\#Queries} & \textbf{Edit Dist}    \\
\midrule 
\multirow{ 3}{*}{\textbf{ATIS}} & \multirow{3}{*}{96.7}  & CWBA(ours) & \textbf{9.3} & \textbf{90.0} & \textbf{2.4} & \textbf{2.0}\\
 & & DeepWordBug & 27.4 & 71.2 & 58.8 & 3.4 \\
 & & Zéroe & 15.2 & 84.3 & 23.4 & 3.8 \\
\midrule 
 \multirow{ 3}{*}{\textbf{SNIPS}} & \multirow{3}{*}{95.8} & CWBA(ours) & \textbf{15.3} & \textbf{86.3} & \textbf{2.6} & \textbf{1.9}\\
 & &DeepWordBug  & 29.5 & 70.1 & 43.9 & 3.5  \\
 & &Zéroe  & 25.3 & 73.5 & 56.1 & 3.0 \\
 \midrule 
\multirow{ 3}{*}{\textbf{CONLL2003}} & \multirow{3}{*}{93.2}  &CWBA(ours) & \textbf{14.4} & \textbf{87.5} & \textbf{3.0} & \textbf{2.9}\\
 & &DeepWordBug & 38.4 & 62.3 & 47.9 & 7.5 \\
 & &Zéroe &33.5 & 66.5 & 49.8 & 7.1 \\
 \midrule 
  \multirow{ 3}{*}{\textbf{OntoNotes}} & \multirow{3}{*}{87.6} &CWBA(ours) & \textbf{5.8} & \textbf{96.2} & \textbf{2.1} & \textbf{2.1}\\
 & &DeepWordBug & 26.3 & 73.9 & 26.1 & 3.5\\
 & &Zéroe & 18.4 & 83.4 & 31.2 & 3.1\\
\bottomrule
\end{tabular}}
\vspace{-2mm}
\caption{Attack result on token classification datatsets with finetuned BERT classifiers.}
\label{tab:token}
\end{table}

Overall, our {\modelname} outperforms the previous baselines in terms of adversarial accuracy and edit distance on all datasets. More specifically,  compared to the previous best methods, our {\modelname} could further reduce the model's accuracy and the edit distance by 0.3 and 6.0 on average respectively. Meanwhile, our required query number is similar to the GBDA model and far less than other black-box methods.  Also, our {\modelname} outperforms the character-level attack method DeepWordBug by a large gap (20.0 adversarial accuracy on average), which demonstrates the advantages of the white-box attack. \\

\noindent\textbf{Token-level attacks.} The attack performance for the BERT classifier towards four token classification datasets is shown in Table \ref{tab:token}.  Similarly, we randomly select 1000 inputs from the test set as attack targets.  For SNIPs and ATIS where the size of the test set is below 1000, the whole test set is selected.
The success rate is the percentage of entities whose predictions are changed after the attack. And we report the average edit distance for entities. 

In general, our {\modelname} surpasses the previous black-box attack methods by a large margin. To be specific,  our {\modelname} could achieve a higher attack success rate (13.08\% on average) than the previous best method with a smaller edit distance (1.9 on average).  Furthermore, our required query number is much less than the previous methods (32.8 on average).  These experimental results demonstrate the effectiveness of our white-box attack method for the token classification task.\\

\noindent\textbf{Attack different transformer models.} To illustrate the generalizability of our {\modelname} towards transformer models, we report the attack result on three different transformer-based models in Table \ref{tab:plm}. We select two benchmarks for sentence and token classification respectively, where all dataset settings are the same as above. 

\begin{table}[bt!]
\centering
\resizebox{\linewidth}{!}{
\begin{tabular}{lccccc}
\toprule
\textbf{Architecture} & \textbf{Datasets}  & \textbf{Clean Acc.(F1.)}  & \textbf{Adv.Acc.(F1.)} & \textbf{\#Queries} & \textbf{Edit Dist} \\
 \midrule 
 \multirow{4}{*}{\textbf{ALBERT}} & \multirow{ 1}{*}{\textbf{ATIS}}&96.3  & \textbf{2.2} &  \textbf{2.1} & \textbf{2.0}  \\
& \multirow{ 1}{*}{\textbf{OntoNotes}} &87.3  &  \textbf{0} & \textbf{1.8}& \textbf{1.6}  \\
 \cmidrule{2-6}
 & \multirow{1}{*}{\textbf{AG News}} & 93.8  & \textbf{2.8} & \textbf{5.6} & \textbf{14.5}  \\
& \multirow{1}{*}{\textbf{Yelp}} & 96.9  & \textbf{3.7} & \textbf{5.5} & \textbf{15.6} \\
 \midrule 
\multirow{4}{*}{\textbf{XLNet}}&\multirow{1}{*}{\textbf{ATIS}}& 96.2  & \underline{16.6} & 4.8 & \underline{3.2}   \\
&\multirow{1}{*}{\textbf{OntoNotes}} &87.6 & 7.1 & 3.0 & \underline{2.5} \\
 \cmidrule{2-6}
 & \multirow{1}{*}{\textbf{AG News}} &94.6  & 4.6 & 5.7 & 16.8  \\
&\multirow{1}{*}{\textbf{Yelp}} &  96.5  & 5.5 & 6.2 & 19.1 \\
 
 \midrule 
 \multirow{4}{*}{\textbf{RoBERTa}}& \multirow{1}{*}{\textbf{ATIS}}& 96.6  & 13.0 & \underline{5.8} & 2.5  \\
& \multirow{1}{*}{\textbf{OntoNotes}} & 87.7  & \underline{17.0} & \underline{4.3} & 2.2  \\
 \cmidrule{2-6}
& \multirow{1}{*}{\textbf{AG News}} & 94.7  & \underline{7.5} & \underline{7.2} & \underline{20.1}   \\
& \multirow{1}{*}{\textbf{Yelp}} & 97.2  & \underline{8.4} & \underline{6.9} & \underline{21.1} \\

\bottomrule
\end{tabular}}
\vspace{-1mm}
\caption{Attack result on three different transformer based pretrained language models.}
\label{tab:plm}
\end{table}

It could be observed that our {\modelname} has an excellent attack performance on these transformer models. Meanwhile, we found that ALBERT is more vulnerable to attacks than other transformer models while even the best-performing models RoBERTa and XLNet are easy to be attacked. These experimental results illustrate that the current pre-training models are vulnerable to character modifications.

\noindent\textbf{Ablation study.} We conduct ablation studies to show the effectiveness of different modules of {\modelname} to the overall attack performance.  The experiments are performed on CONLL and AG News datasets with the BERT classifier. {\modelname} \texttt{w random word selection} random selects target words instead of searching by gradient.  {\modelname} \texttt{w/o visual constraint} and \texttt{w/o length constraint} removes the visual and length constraint respectively.  {\modelname} \texttt{w/o visual \& length constraint} only keeps the adversarial attack object without any constraint.

\begin{table}[bt!]

\resizebox{0.99\linewidth}{!}{
\begin{tabular}{llccc}
\toprule
\multicolumn{1}{c}{\textbf{Dataset}} & \multicolumn{1}{c}{\textbf{Technique}} & \multicolumn{1}{c}{\textbf{Adv Acc.(F1.)}} & \multicolumn{1}{c}{\textbf{\#Queries}} & \multicolumn{1}{c}{\textbf{Edit Dist}}  \\
\midrule 
\multirow{5}{*}{\textbf{CONLL}} &\modelname & 9.1 & 4.5 & \textbf{2.7}\\
& w random word selection  & 59.4  & 7.3   & 2.9  \\
& w/o visual constraint & 8.7   & 3.8  & 3.3 \\
& w/o length constraint & 9.3  &  4.2   &  3.2  \\
& w/o visual \& length constraint & \textbf{6.2}  & \textbf{3.9}  &  4.7  \\
\midrule 
\multirow{5}{*}{\textbf{AG News}} &\modelname & 3.2 & 6.1 & \textbf{17.3}\\
& w random word selection  & 54.2  & 13.4   & 26.7  \\
& w/o visual constraint & 2.7  & 5.6  & 25.6 \\
& w/o length constraint & 3.0  & 5.8   &  26.7  \\
& w/o visual \& length constraint & \textbf{2.4}  & \textbf{4.8}   & 37.8  \\
\bottomrule
\end{tabular}}
\caption{Ablation study of how different modules contribute to the attack performance on BERT classifier.}
\label{tab:ablation}
\end{table}

\begin{figure}[t!]
    \centering
    \includegraphics[width=0.49\linewidth]{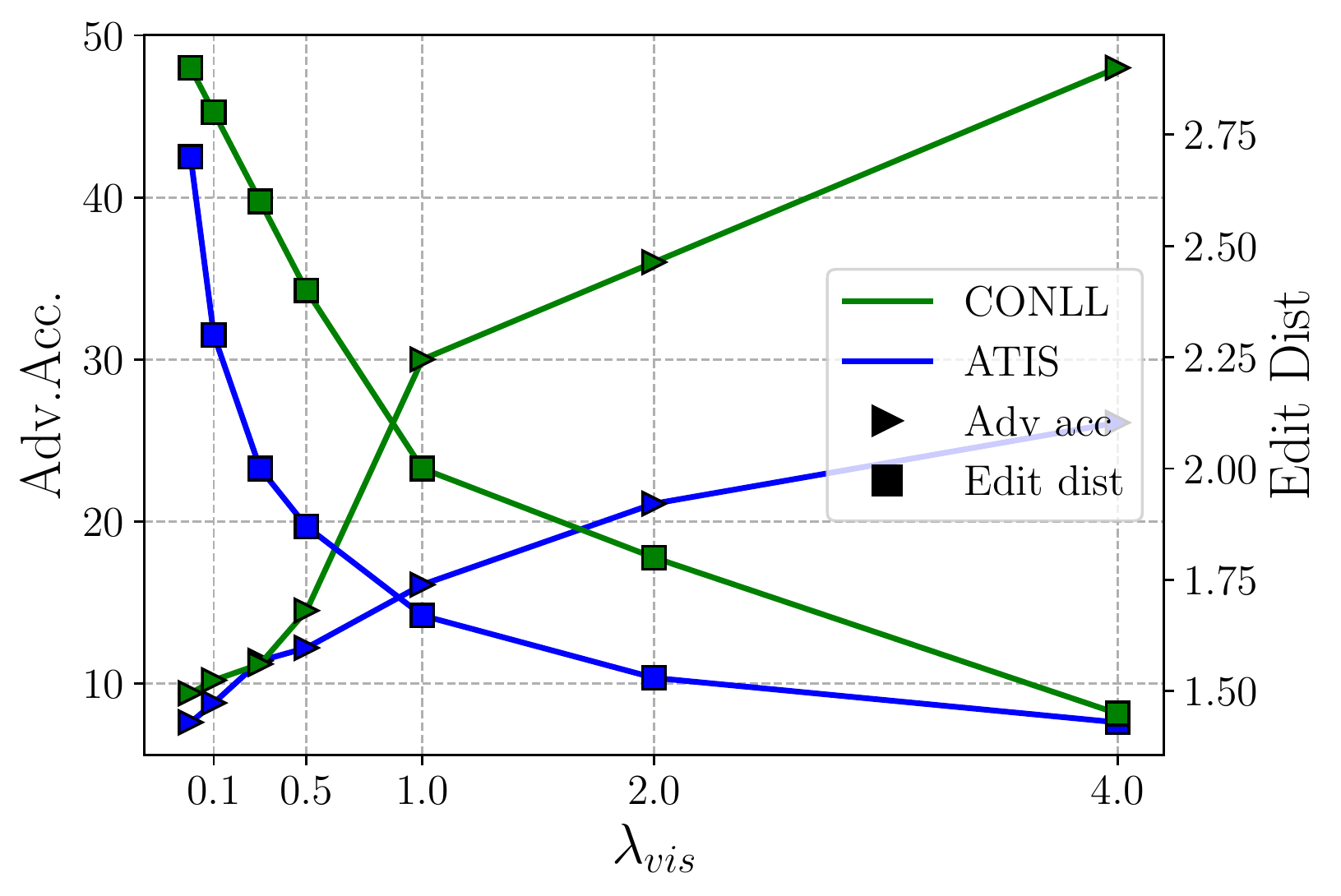}
    \includegraphics[width=0.49\linewidth]{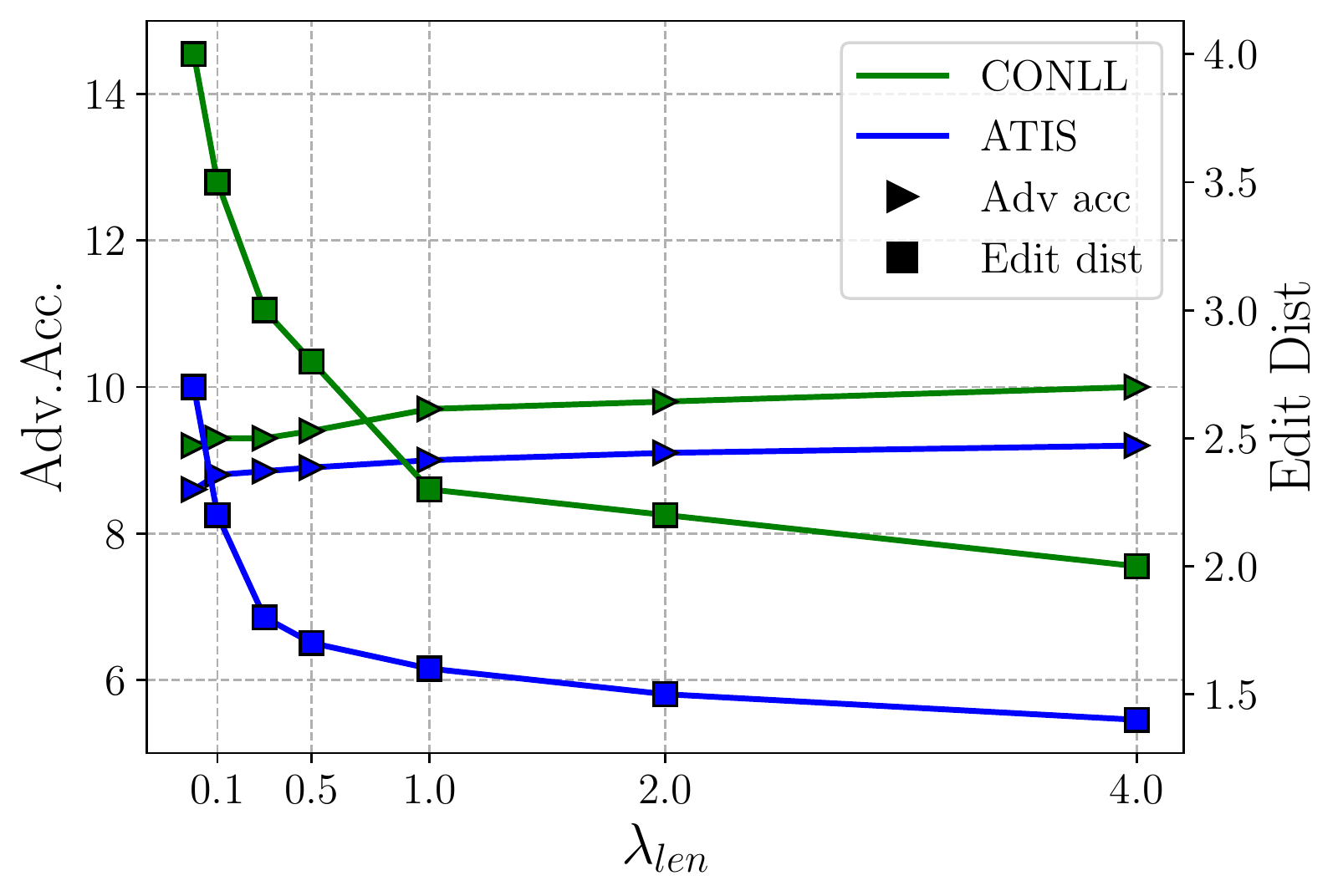}
    \caption{Adversarial accuracy and edit distance during the growth of $\lambda_{vis}$(left) and $\lambda_{len}$(right).}
    \label{fig:trend}
\end{figure}

Our observations from the experimental results in Table \ref{tab:ablation} are as follows: (1) The target word selection module has a huge impact on the attack success rate. (2) The visual constraint could significantly reduce the edit distance while sacrificing a little attack performance. Further analysis would help us balance the attack success rate and edit distance. (3) The length constraint has little effect on the attack performance but could effectively reduce the edit distance. In general, all our modules have positive effects on the attack performance.

\subsection{Analysis}
\label{sec: ana}

\noindent\textbf{Effectiveness of visual and length constraint.} To further investigate how the visual and length constraint contributes to the attack performance, we visualize the changing trend of adversarial accuracy and edit distance when $\lambda_{vis}$ and $\lambda_{len}$ grow from zero in Figure \ref{fig:trend}.  When tuning one hyperparameter, the other hyperparameter would be set to 0.1. These experiments are performed on two token classification datasets: CONLL and ATIS.

It can be seen that with $\lambda_{vis}$ increases, the edit distance decreases while the attack performance drops (the accuracy after the attack increases). Meanwhile, with the increase of $\lambda_{len}$,  the attack performance is almost unchanged while the edit distance still reduces. So we conclude that $\lambda_{vis}$ influences more on the attack success rate than $\lambda_{len}$,  which may be because the transformer-based language models are robust to visual similarity changes to some extent. These experiment results inspire us to adopt a larger  $\lambda_{len}$ than $\lambda_{vis}$. \\

\begin{table}[bt!]
\centering
\resizebox{\linewidth}{!}{
\begin{tabular}{lcccc}
\toprule
\multicolumn{1}{c}{\textbf{Dataset}} & \multicolumn{1}{c}{\textbf{Ori.Acc.(F1.)}} & \multicolumn{1}{c}{\textbf{Adv.Acc.(F1.)}}  & \multicolumn{1}{c}{\textbf{Identification}} & \multicolumn{1}{c}{\textbf{Correction}}  \\
\midrule 
\multirow{1}{*}{\textbf{ATIS}} &  97.2 & 94.5 & 99.1 & 93.3\\
\multirow{1}{*}{\textbf{SNIPS}}&   96.3 & 92.1 & 97.8 & 90.5  \\
\midrule 
\multirow{1}{*}{\textbf{Yelp}} & 91.2 & 82.2 & 92.4 & 83.1\\
\multirow{1}{*}{\textbf{AG News}} & 90.3 & 80.8 & 93.4 & 82.4 \\
\bottomrule
\end{tabular}}
\caption{Human evaluation for the ability to identify and correct adversarial examples.}
\label{tab:human}
\end{table} 

\begin{table}[bt!]

\resizebox{0.99\linewidth}{!}{
\begin{tabular}{llccc}
\toprule
\multicolumn{1}{c}{\textbf{Dataset}} & \multicolumn{1}{c}{\textbf{Technique}} & \multicolumn{1}{c}{\textbf{Adv Acc.(F1.)}} & \multicolumn{1}{c}{\textbf{\#Queries}} & \multicolumn{1}{c}{\textbf{Edit Dist}}  \\
\midrule 
\multirow{6}{*}{\textbf{ATIS}} & \modelname & \textbf{9.3} & 2.4 & \textbf{2.0}\\
& \modelname + Adv.Training & 58.5  & 9.3   & 4.8  \\
 \cmidrule{2-5}
& DeepWordBug  & \textbf{27.4}  & 58.8   & \textbf{3.4} \\
& DeepWordBug  + Adv.Training & 44.8  & 38.4   & 4.8  \\
 \cmidrule{2-5}
& Zéroe  & \textbf{15.2}  & 23.4   & 3.8  \\
& Zéroe  + Adv.Training & 56.3  & 19.5   & \textbf{3.2}  \\
\midrule 
\multirow{4}{*}{\textbf{AG News}} &\modelname & \textbf{3.2} & 6.1 & \textbf{17.3}\\
& \modelname + Adv.Training  & 60.3  & 12.4   & 31.2  \\
 \cmidrule{2-5}
& DeepWordBug   & \textbf{23.7}  & 319   & \textbf{23.4}  \\
& DeepWordBug + Adv.Training  & 57.9  & 412   & 33.5  \\
\bottomrule
\end{tabular}}
\caption{Model robustness improvement after adversarial learning with the generated adversarial examples.}
\label{tab:adversarial}
\end{table}

\begin{table*}[bt!]
\resizebox{0.99\linewidth}{!}{
\begin{tabular}{llll}
\toprule
\multicolumn{3}{l}{\textbf{Dataset}} & \multicolumn{1}{l}{\textbf{Label}} \\
\midrule 
 \multirow{4}{*}{\textbf{AG News}}& \multirow{2}{*}{Ori} & fund pessimism grows new york ( {\color{blue} cnn} / money ) - money managers are growing more pessimistic about the  {\color{blue} economy}, & \multirow{2}{*}{{\color{blue} Business}}  \\ &&corporate profits and us stock market returns, according to a monthly survey by merrill lynch released tuesday. & \\
  \cmidrule{2-4}
 & \multirow{2}{*}{Adv} & fund pessimism grows new york ( {\color{red} cn\cjRL{t}} / money ) - money managers are growing more pessimistic about the  {\color{red} econ0my}, & \multirow{2}{*}{{\color{red} Sci/Tech}}\\  &&corporate profits and us stock market returns, according to a monthly survey by merrill lynch released tuesday. &  \\
\midrule 
 \multirow{4}{*}{\textbf{Yelp}}& \multirow{2}{*}{Ori} & seattle may have just won the 2014 super bowl, but the steelers still {\color{blue} rock} with six {\color{blue} rings}, {\color{blue}baby}!!! just {\color{blue} stating} what all   & \multirow{2}{*}{{\color{blue} Positive}}  \\ &&  steeler fans know : a steel dynasty is still unmatched no matter what team claims the title of current super bowl champs. & \\
  \cmidrule{2-4}
 & \multirow{2}{*}{Adv} & seattle may have just won the 2014 super bowl, but the steelers still {\color{red} robk} with six {\color{red} riigs}, {\color{red}bacy}!!! just {\color{red} staging} what all  & \multirow{2}{*}{{\color{red} Negative}}\\  && steeler fans know : a steel dynasty is still unmatched no matter what team claims the title of current super bowl champs. &  \\
\midrule 
 \multirow{2}{*}{\textbf{OntoNotes}}& Ori & So far, the \underline{{\color{blue} French}} have failed to win enough broad - based support to prevail.  \qquad  \qquad   Prediction span:  \underline{French}  &  {\color{blue} \underline{NORP}}     \\
  \cmidrule{2-4}
 & Adv & So far, the \underline{{\color{red}Fr€nch}} have failed to win enough broad - based support to prevail.   \qquad  \qquad   Prediction span:  \underline{Fr€nch}  &  {\color{red} \underline{ORG}}   \\ 
\bottomrule
\end{tabular}}
\caption{The generated adversarial examples.  The origin  {\color{blue} label} is the correct prediction and the adversarial {\color{red}label} is adverse prediction. The first two examples are from sentence classification task while the third case is from the token classification task. The \underline{target tokens} and \underline{labels} for token classification are underlined. }
\label{tab:case}
\end{table*}

\noindent\textbf{Human evaluation.} To examine whether the attacked text preserves its original label, we set up human evaluations to measure the quality of the generated text.  We first ask human judges to make predictions on both the original and attacked texts.  Then we ask them to identify and correct the modified words in the attacked text.  The evaluations are conducted on 100 selected sentences from ATIS, SNIPs DBPedia, and AG News datasets respectively. For each dataset, we ask three human evaluators to measure the quality of examples.

The human evaluation results are presented in Table \ref{tab:human}. And we could observe that: (1) Human judges could predict most of the attacked text correctly, which demonstrates our generated examples are label-preserving. (2) Although most of the modified words could be identified, most of these misspelled words could be corrected by humans, which does not hinder understanding. (3) Perturbations on the words in sentence classification datasets are harder to identify and correct than that in token classification datasets because of the larger edit distance.  \\

\noindent\textbf{Adversarial training.} To further explore whether the adversarial examples could help improve the model's robustness, we perform an adversarial training experiment by training the model with the combination of the original and the adversarial examples.  Specifically, the adversarial examples are selected from the attacked texts of the ATIS and DBPedia training sets.  Furthermore, we  compare the attack results towards the original model and the adversarially trained model using our {\modelname} and other character-level attack methods.

We present the adversarial training results in Table \ref{tab:adversarial}. Overall, the robustness of our model towards character-level attacks   improves tremendously after the adversarial training. Specifically, the attack success rate of our {\modelname} decreases drastically (53.15 on average), while the required editing distance and the number of queries become much larger (6.6 and 8.4 on average). Similarly, the attack performance of other character-level attack methods drops severely in all metrics. These experimental results indicate that our generated texts could preserve the origin label.\\

\noindent\textbf{Tokenization analysis.} Our {\modelname} works on the tokenized subtokens. However, the adversarial texts are re-tokenized before being fed to the model, where the re-tokenization result may not be the same set of origin subtokens. For example, the subtokens \textit{bo-}, \textit{sl-} and \textit{on} would be re-encoded into \textit{bos-} and \textit{lon}. We further analyze how tokenization inconsistency affects attack performance.

In practice, we observe that the tokenization inconsistency doesn't impact the attack performance by much. Specifically,  31.2\% words are not re-tokenized to the same subtoken set but only lead to 3\% attack failures, which indicates the re-tokenized examples are still adversarial. Similar observations are reported in previous works \cite{guo-etal-2021-gradient}. 

We further analyze the reason of attack failures and find 65\% of failed examples are not optimized well and tokenization inconsistency leads to 35\% attack failures, which indicates that tokenization inconsistency has a limited impact on our method. \\

\noindent\textbf{Case study.}  To intuitively show the effectiveness of {\modelname},  we select three cases to compare the original and adversarial texts.  These cases are sampled from AG News, Yelp (sentence classification), and OntoNotes (token classification) datasets.

As seen in Table \ref{tab:case}, the generated adversarial sentences are semantically consistent with their original texts, while the target model makes incorrect predictions.  Meanwhile, we could observe that many words are visually similar during the attack (e.g. \textit{cnn} and \textit{cn\cjRL{t}}), which shows the effectiveness of our visual constraint.  The number of words to be attacked for sentence-level tasks is larger than the token-level tasks, and the concrete number is also uncertain (two in the first case and four in the second case). For token classification tasks like NER, the attacked words are usually the entities.

\section{Conclusions}

In this paper, we propose {\modelname}, the first character-level white-box attack method for transformer models. We substitute the attachable subtokens to achieve character modification. The Gumbel-Softmax technique is adopted to allow gradient propagation. Meanwhile, the visual and length constraint help preserve the semantics of adversarial text. Experiments on both sentence-level and token-level tasks on various transformer models demonstrate the effectiveness of our method.  

\section*{Acknowledgement}

The work was supported by the National Key Research and Development Program of China (No.2019YFB1704003), the National Nature Science Foundation of China (No.62021002), Tsinghua BNRist and Beijing Key Laboratory of Industrial Big Data System and Application.

\section*{Limitations}
The major limitation of {\modelname} has been discussed in the tokenization analysis part in section \ref{sec: ana}: the generated words may be re-tokenized into different subtokens.  Although most examples are still adversarial, it introduces uncontrollability.  We hope future works could introduce more constraints to alleviate this problem.

Also, we achieve character modification by subword substitution, but not all combinations of characters exist in the vocabulary. Therefore, the effect of our attack method depends on the size of the vocabulary.


\bibliography{anthology,custom}
\bibliographystyle{acl_natbib}

\appendix



\section{Attachable subwords analysis}

\begin{table}[h]
\centering
\resizebox{0.95\linewidth}{!}{
\begin{tabular}{llll}
\toprule
\multicolumn{1}{c}{\textbf{Subwords len}} & \multicolumn{1}{c}{\textbf{Num in Vocab}} & \multicolumn{1}{c}{\textbf{Potential Num}} & \multicolumn{1}{c}{\textbf{Ratio}}\\
\midrule 
1 & 26 & 26 & 1\\
2 & 438 & 676 & 0.65 \\
3 & 1438 & 17576 & 0.08 \\
4 & 1573 & 456976 & 0.03 \\
5 & 695 & 11876696 & 0.5$\times$10-5 \\
\bottomrule
\end{tabular}}
\caption{The subword number of different lengths in the vocabulary.}
\label{tab:subwords}
\end{table}

To better illustrate the principles of adversarial tokenization, we list the statistics for the number of attachable subwords with different lengths in Table \ref{tab:subwords}.  We can see that with the length increases, the proportion of subwords in vocabulary among all potential subwords with the same length is getting smaller.  For example, all the 26 attachable subwords with length 1 (\textit{\#a} - \textit{\#z}) are in the vocabulary list, but some subwords with length 2 (\textit{\#rz}) doesn't. 

Based on these observations, we conclude that the longer subtokens are more difficult to combine with characters to form new subwords, which is the principle of the adversarial tokenization module.

\section{Details of baselines}

\noindent\textbf{Token classification task.} Token classification is a natural language understanding task in which a label is assigned to some tokens in the text. Some popular token classification subtasks are Named Entity Recognition (NER) \cite{zhang2021crowdsourcing, shen2021locate, ijcai2021p0542, shen2022parallel, zhang2022identifying} and Slot Filling \cite{chen2019bert, zhang2017position}.

Since the token classification model generates a label for each token, the token itself cannot be replaced, and the structure of the sentence also cannot be modified. Therefore, neither sentence-level \cite{xu2020rewriting} nor word-level attacks \cite{eger-benz-2020-hero} can be applied to the token classification task, only character-level attacks are available in this scenario.

The current character-level attack methods can be divided into two categories. The first class of methods performs a white-box attack against a model taking characters as input. The most representative methods is \textbf{HotFlip} \cite{ebrahimi-etal-2018-hotflip}. Other works are mainly variants of HotFlip \cite{ebrahimi-etal-2018-adversarial,gil2019white}.  Another class of methods performs black-box attacks on the model, where main representative methods are  \textbf{DeepWordBug} \cite{gao2018black} and \textbf{Zéroe} \cite{eger-benz-2020-hero}. These methods do not require the input to the model to be characters.  Since our approach attacks models with word-level input, we mainly take DeepWordBug and Zéroe as our baselines.

\noindent\textbf{Sentence classification task} 
Sentence classification is one of the simplest NLP tasks in the Natural Language Processing field that have a wide range of applications including sentiment analysis \cite{tang2015effective, tang2014learning} and relation extraction \cite{hu2021semi, hu2021gradient, hu2020selfore, liu2022hierarchical}.

Since the sentence classification method outputs a label for the whole sentence, both word-level attacks and character-level attacks can be performed on it.  Word level attack method is the most widely used text attack method, which can be classified into the black-box method and white-box method.  The classical black-box approach mainly uses a rule-based approach to do synonym replacement, of which the most representative is \textbf{TextFooler} \cite{jin2020bert}. Recent methods like \textbf{BERT-Attack} \cite{li-etal-2020-bert-attack} and \textbf{BAE} \cite{garg-ramakrishnan-2020-bae} replace words in context with the help of semantic information from the pre-trained model \cite{devlin-etal-2019-bert}.  The representative work of white-box attack is \textbf{GBDA} \cite{guo-etal-2021-gradient}, which performs token replacement under the gradient guidance.  In this work, we adopt token-level attack methods TextFooler, BAE, BERT-Attack, GBDA and  character-level attack method DeepWordBug as our baselines.



\end{document}